\documentclass[sn-mathphys-num]{sn-jnl}

\usepackage{graphicx}%
\usepackage{multirow}%
\usepackage{amsmath,amssymb,amsfonts}%
\usepackage{amsthm}%
\usepackage{mathrsfs}%
\usepackage[title]{appendix}%
\usepackage{xcolor}%
\usepackage{textcomp}%
\usepackage{manyfoot}%
\usepackage{booktabs}%
\usepackage{algorithm}%
\usepackage{algorithmicx}%
\usepackage{algpseudocode}%
\usepackage{listings}%
\usepackage{graphicx}
\usepackage{subfigure}
\usepackage{bm}

\theoremstyle{thmstyleone}%
\DeclareMathOperator{\tr}{tr}
\DeclareMathOperator*{\argmin}{arg\,min}

\theoremstyle{thmstyletwo}%

\theoremstyle{thmstylethree}%

\begin{document}

\title[Article Title]{Information-Theoretic Greedy Layer-wise Training for Traffic Sign Recognition}


\author[1]{\fnm{Shuyan} \sur{Lyu}}\email{2024005980@link.tyut.edu.cn}
\equalcont{These authors contributed equally to this work.}

\author[2]{\fnm{Zhanzimo} \sur{Wu}}\email{zcahwuf@ucl.ac.uk}
\equalcont{These authors contributed equally to this work.}

\author*[3]{\fnm{Junliang} \sur{Du}}\email{jldu@acm.org}

\affil[1]{\orgdiv{School of Software}, \orgname{Taiyuan University of Technology}, \orgaddress{\city{Taiyuan}, \postcode{100190}, \state{Shanxi}, \country{China}}}

\affil[2]{\orgdiv{Department of Mathematics and Statistical Science}, \orgname{University College London}, \orgaddress{\city{London}, \postcode{WC1E 6BT}, \country{United Kingdom}}}

\affil*[3]{\orgdiv{AI Institute}, \orgname{Shanghai Jiao Tong University}, \orgaddress{\city{Shanghai}, \postcode{200240}, \country{China}}}


\abstract{Modern deep neural networks (DNNs) are typically trained with a global cross-entropy loss in a supervised end-to-end manner: neurons need to store their outgoing weights; training alternates between a forward pass (computation) and a top-down backward pass (learning) which is biologically implausible. Alternatively, greedy layer-wise training eliminates the need for cross-entropy loss and backpropagation. By avoiding the computation of intermediate gradients and the storage of intermediate outputs, it reduces memory usage and helps mitigate issues such as vanishing or exploding gradients.
However, most existing layer-wise training approaches have been evaluated only on relatively small datasets with simple deep architectures. 
In this paper, we first systematically analyze the training dynamics of popular convolutional neural networks (CNNs) trained by stochastic gradient descent (SGD) through an information-theoretic lens. Our findings reveal that networks converge layer-by-layer from bottom to top and that the flow of information adheres to a Markov information bottleneck principle. Building on these observations, we propose a novel layer-wise training approach based on the recently developed deterministic information bottleneck (DIB) and the matrix-based R\'enyi's $\alpha$-order entropy functional. Specifically, each layer is trained jointly with an auxiliary classifier that connects directly to the output layer, enabling the learning of minimal sufficient task-relevant representations. We empirically validate the effectiveness of our training procedure on CIFAR-10 and CIFAR-100 using modern deep CNNs and further demonstrate its applicability to a practical task involving traffic sign recognition. Our approach not only outperforms existing layer-wise training baselines but also achieves performance comparable to SGD.}




\maketitle

\section{Introduction}\label{sec1}

\begin{figure}[h!]
\centering
\includegraphics[width=1\textwidth]{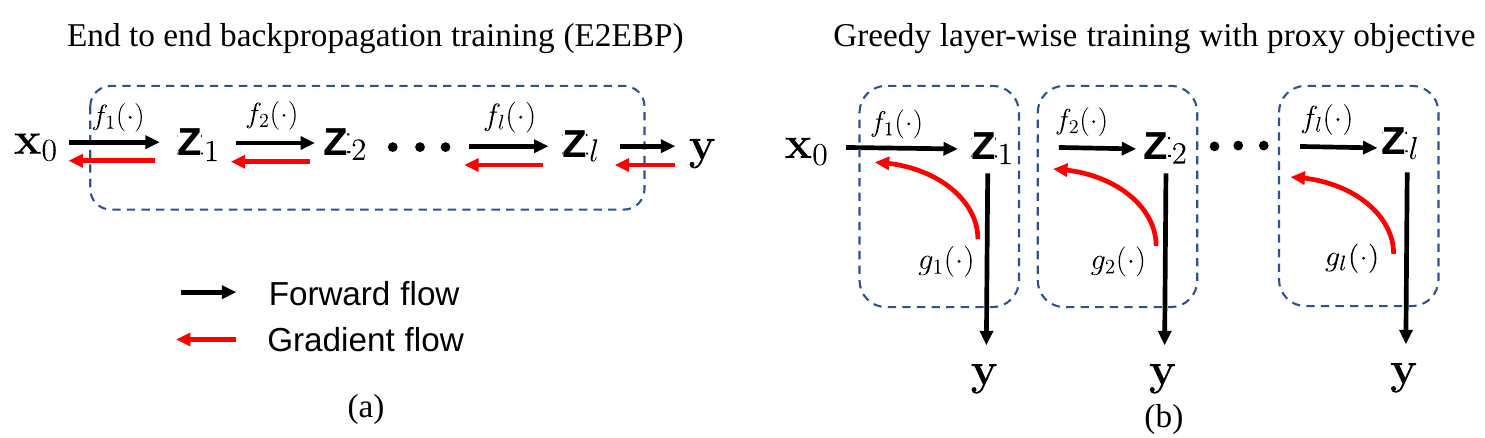}
\caption{Diagram of (a) E2EBP and (b) greedy layer-wise training with a local proxy objective. $f_l(\cdot)$ represents the $l$-th layer of the whole model,  $g_l(\cdot)$ denotes an auxiliary classifier. $g_l(\cdot)$ was trained together with $f_l(\cdot)$ to minimize a local
proxy objective and was discarded at test time.}\label{figure_1}
\end{figure}

End-to-end back-propagation (E2EBP) (see Fig.~\ref{figure_1}(a)) by stochastic gradient descent (SGD) or its variants such as ADAM~\cite{Adam2015} is a dominating strategy to train modern deep neural networks (DNNs). Typically, the gradients of the cross-entropy loss with respect to the network weights are computed at the final layer and propagated backward through the network, layer by layer, to update the weights.
However, E2EBP suffers from several limitations. First, it lacks biological plausibility, as the human brain does not appear to have a centralized mechanism for computing and distributing a global error signal across all neurons~\cite{pogodin2020kernelized}. Second, it is prone to issues such as vanishing or exploding gradients~\cite{duan2021training,hinton2022forward}. Additionally, backpropagation requires all hidden layer weights to be retained in memory until the forward and backward passes are complete, leading to significant memory usage. This backward locking phenomenon prevents memory reuse, which poses a critical limitation for resource-constrained or on-device settings~\cite{karkar2024module}.

To mitigate the above limitations, there have been several attempts to design layer-wise learning approaches without a global cross-entropy loss or backpropagation. Specifically, a DNN is first divided into several gradient-isolated layers or modules and is then trained successively based on local supervision signals~\cite{karkar2024module}. Well-known examples of local supervision signals include target label information~\cite{belilovsky2019greedy}, synthetic gradients~\cite{jaderberg2017decoupled}, and information-theoretic metrics~\cite{ma2020hsic}. The layer-wise training approach is more computationally efficient than E2EBP and mitigates the risk of gradient vanishing or exploding, as forward and backward propagation are performed only within a single layer or module. Furthermore, layer-wise training is considered more biologically plausible, reflecting the modular nature of the brain, which predominantly learns from local signals~\cite{crick1989recent}.

Previous studies on layer-wise training can be broadly categorized into the following approaches: Target Propagation~\cite{bengio2014auto,lee2015difference, meulemans2020theoretical}, Synthetic Gradients~\cite{jaderberg2017decoupled,czarnecki2017understanding,lansdell2019learning}, and Proxy Objective~\cite{belilovsky2019greedy, nokland2019training, wu2022deep,DuanYCP20}. Neither target propagation-based nor synthetic gradients-based approaches are scalable to challenging benchmark datasets or modern deep models, such as VGG~\cite{simonyan2014very} and ResNet~\cite{he2016deep}. In contrast, the proxy objective achieves the best empirical performance among all alternatives to E2EBP training. A significant class of proxy objective-based methods directly integrates target labels into the design of a local objective for latent representations, encouraging the latent outputs to closely align with the target labels, as illustrated in Fig.~\ref{figure_1}(b). However, these local proxy objectives often lack robust theoretical or empirical validation. Furthermore, their performance is typically assessed only on small-scale benchmark datasets (e.g., MNIST) using simple deep architectures (e.g., fully connected neural networks). No prior approach can yet match the performance of E2EBP, making them less practical in comparison~\cite{sakamoto2024end}.
Consequently, their practical effectiveness on real-world applications involving more complex deep architectures remains unclear.


In this work, we first conduct a systematic analysis of the learning dynamics of DNNs trained with E2EBP through an information-theoretic lens. Our analysis demonstrates that DNNs converge progressively from shallower layers to deeper layers and that the information flow within feed-forward deep architectures follows the Markov information bottleneck principle~\cite{nguyen2019markov}, as discussed in detail in Section~\ref{sec:analysis}.
These observations imply that even when a DNN is trained using E2EBP, its internal training dynamics inherently follow a layer-wise progression. In other words, E2EBP implicitly performs layer-wise training. This insight helps bridge the conceptual gap between E2EBP and explicit layer-wise training approaches. Building on our observations, we propose a novel greedy layer-wise training approach inspired by the recently developed deterministic Information Bottleneck (DIB)~\cite{strouse2017deterministic,yu2021deep}. We evaluate its performance using ResNet on CIFAR-100 and further demonstrate its applicability by extending the approach to a real-world task of traffic sign detection.


To summarize, our main contributions include: 
\begin{itemize}
    \item We conduct a systematic analysis to bridge the conceptual gap between E2EBP and explicit layer-wise training. Our empirical observations also reveal a more promising local proxy objective that differs from previous proposals by explicitly introducing an information compression term.
    \item Based on our observations, we propose a novel greedy layer-wise training approach called Greedy Deterministic Information Bottleneck (Greedy-DIB). Our Greedy-DIB outperforms existing state-of-the-art (SOTA) layer-wise training approaches and achieves performance on par with SGD. Note that, previous studies typically evaluated these approaches on benchmark classification tasks (e.g., MNIST) using MLPs, while our Greedy-DIB demonstrates strong generalization on CIFAR-100 with ResNet-18.
    \item To further demonstrate the effectiveness of our approach, we also test its performance on a real-world traffic sign recognition task, which involves both object classification and bounding box regression (i.e., a two-head neural architecture). To the best of our knowledge, we are among the first to apply a local loss-based CNN for traffic sign recognition.
\end{itemize}

\section{Related Work}
\subsection{Layer-wise Training with Proxy objective}
Recently, proxy objective methods have become increasingly popular.
Greedy LS~\cite{belilovsky2019greedy} and Greedy LE~\cite{nokland2019training} add an auxiliary model to each single layer during the training. Specifically, Greedy LS utilizes only the cross-entropy loss between the output of the auxiliary model and labels. In contrast, Greedy LE combines cross-entropy loss with a cosine similarity regularization between the output of the auxiliary model and the label. The objective of Neural Indicator Kernel (NIK)~\cite{wu2022deep} is essentially the normalized Hilbert-Schmidt Independence Criterion (nHSIC)~\cite{gretton2005measuring}, which measures the dependency between the latent representation and the label without relying on any auxiliary models. However, these objectives are also prone to learning redundant information in each layer, as the designed learning objective includes only a prediction term that evaluates the latent representation's predictive power with respect to the label. As a result, the learned representation often captures noisy or redundant information from the input $X$ that is irrelevant to downstream tasks. This partially explains why no prior approach has been able to match the performance of E2EBP, rendering them less practical.


The most relevant work to ours is the HSIC bottleneck~\cite{ma2020hsic}, which also employs the Information Bottleneck (IB) principle~\cite{tishby2000information} to train deep networks without relying on error backpropagation. Unlike the HSIC bottleneck, our objective is based on the deterministic IB framework~\cite{strouse2017deterministic}, which, as we will demonstrate in later sections, offers more theoretical justifications. Additionally, instead of utilizing nHSIC, we adopt the recently proposed matrix-based R{\'e}nyi's $\alpha$-order entropy functional~\cite{giraldo2014measures,yu2019multivariate}, which simplifies the estimation of the mutual information term. Moreover, our approach significantly outperforms the HSIC Bottleneck, as will be demonstrated in Section~\ref{sec:experiments}.
We also observed that \cite{DuanYCP20} utilizes kernel alignment~\cite{cristianini2001kernel} as a proxy objective. However, its effectiveness has only been evaluated on kernel networks, a specialized type of neural network where neurons are replaced with kernel machines. The objective
function for the above approaches and our proposed method are shown in Table \ref{loss_function}.

\begin{table}[!ht]
\centering
\caption{Existing layer-wise training approaches and their corresponding objective in the $l$-th layer. $x$ refers to the input and $z_l$ refers to the $l$-th hidden layer representation, $y$ is the class label. $g(\cdot)$ represents the auxiliary classifier and $S(\cdot)$ represents cosine similarity.}
\label{loss_function}
\begin{tabular}{cc}
\hline
Method & Objective function\\
\hline
Greedy LS~\cite{belilovsky2019greedy} & $\textmd{CE}(g(z_l);y)$\\
Greedy LE~\cite{nokland2019training} & $(1-\beta)\textmd{CE}(g(z_l);y)+\beta(||S(g(z_l))-S(y)||^2_F)$\\
HSIC bottleneck~\cite{ma2020hsic} & $\textmd{nHSIC}(z_l;x)-\beta\textmd{nHSIC}(z_l;y)$\\
NIK~\cite{wu2022deep} & $-\textmd{nHSIC}(z_l;y)$\\
\hline
\textbf{Ours} & $\beta H(z_l)+\textmd{CE}(g(z_l);y)$\\
\hline
\end{tabular}
\end{table}

\subsection{Understanding Training Dynamics of Deep Neural Networks}
Although deep neural networks have achieved great success, interpreting and understanding their training dynamics remain a challenge. 
Recently, information-theoretic tools have attracted much attention in understanding DNNs. One particular concept is the so-called information plane~\cite{shwartz2017opening}, which depicts the trajectory in $\mathbb{R}^2$ of the mutual information pair $\{I(X; Z), I(Y;Z)\}$ across training epochs, as a lens to analyze dynamics of learning of
DNNs~\cite{yu2021information}. According to~\cite{shwartz2017opening}, there are two training phases in the common stochastic gradient descent (SGD) optimization: an early ``fitting" phase, in which both $I(X;Z)$ and $I(Y;Z)$ increase rapidly, and a later ``compression" phase, in which there is a reversal such that $I(X;Z)$ continually decrease. This work attracted significant attention, culminating in many follow-up works that tested the proclaimed narrative and its accompanying empirical observations.
So far, the ``fitting-and-compression" phenomena of the layered representation $Z$ have been observed in other types of DNNs, including the multilayer perceptrons (e.g.,~\cite{chelombiev2019adaptive,shwartz2017opening}), the deterministic autoencoders (e.g.,~\cite{yu2019understanding}), and the CNNs (e.g.,~\cite{noshad2019scalable,yu2020understanding}).

However, previous literature defines the $x$-axis of the information plane as \( I(X; Z) \). In contrast, we define the $x$-axis of the information plane as the mutual information between the hidden layer representation and its preceding layer, i.e., \( I(Z_i; Z_{i-1}) \). This modification aligns with the essence of layer-wise training, as it reflects the progressive nature of network training, where information is processed and refined layer by layer in a sequential manner.


Another research direction focuses on studying the loss landscapes~\cite{li2018visualizing} during training. A common concept in this area is the so-called sharpness of local minima, which can be quantified using first-order (e.g., Jacobian, Lipschitz constant) or second-order (e.g., Hessian spectrum) sensitivity measures. \cite{KeskarMNST17} observed that DNNs generalize well when they converge to flat minima. \cite{cha2021swad} further improves the generalizability by encouraging the neural networks to find flat minima through the stochastic weight averaging densely. However, the existing sharpness-based metrics fail to provide consistent observations and can be influenced by various training techniques (e.g., adversarial training~\cite{yao2018hessian}, $\ell_2$ regularization~\cite{granziol2020flatness}).

\subsection{Traffic Sign Recognition}
Traffic sign recognition holds significant industrial potential for advanced driver assistance systems and intelligent autonomous vehicles. Generally, traffic signs are categorized based on their functions, such as warning, prohibitory, or mandatory signs. Within each category, signs may be further divided into subclasses that share similar shapes and appearances but differ in specific details (e.g., speed limits of 40, 60, and 80 within the ``speed limit" category). This structure suggests that traffic sign recognition should be approached as a two-phase task: detection followed by classification~\cite{zhu2016traffic}. The detection phase identifies bounding boxes in an image that are likely to contain traffic signs, while the classification phase determines the exact type of sign present (if any). However, several challenges hinder accurate recognition by computer algorithms, including variations in illumination, color degradation, motion blur, cluttered backgrounds, and partial occlusion.

In the deep learning era, convolutional neural networks (CNNs) have been extensively applied to traffic sign detection and classification. \cite{zhu2016traffic} proposed two fully convolutional networks, one for traffic sign detection and the other for simultaneous detection and classification, with both networks sharing the same architecture except for the branches in the final layer. \cite{meng2017detecting} utilized the Single Shot Multibox Detector (SSD) framework with VGG-16 as the backbone for traffic sign detection. \cite{ayachi2020traffic} explored a lightweight MobileNet to balance memory efficiency and accuracy. In recent years, the YOLO family of models has gained popularity in this domain. YOLO~\cite{redmon2016you}, as an initiator, introduced a fully connected layer to directly predict both the object class and bounding box. \cite{wu2020real} proposed an end-to-end framework based on YOLOv3, while \cite{qu2023improved} adopted YOLOv5 for this task.

All these approaches focus on modifying network architectures but overlook the potential of alternative training algorithms. In this paper, we aim to demonstrate that a greedy layer-wise training approach offers a promising alternative to traditional backpropagation, making it particularly suitable for vehicle systems with limited memory resources.

\section{Analyzing End-to-End Backpropagation through an Information-Theoretic Lens} \label{sec:analysis}

In this section, we analyze the learning dynamics of end-to-end training from an information-theoretic perspective. In Section~\ref{sec:two_MI}, We design and visualize the changes in two mutual information matrices over the entire training epochs to demonstrate that, even when a network is trained using E2EBP, its learning dynamics implicitly follow a layer-wise training pattern. Specifically, the convergence of layer representations begins with the shallower layers and gradually progresses to the deeper layers. In Section~\ref{sec:IP}, we introduce a modification to the original information plane proposed in \cite{shwartz2017opening} by replacing the mutual information $I(x;z)$ with the entropy of the latent representation $H(z)$ on the $x$-axis. Our goal is to demonstrate that the latent representation undergoes both a fitting phase and an information compression phase. Previous literature on layer-wise training only models the fitting phase (see Table~\ref{loss_function}), while neglecting the modeling of information compression.




\subsection{Understanding Learning Dynamics by Mutual Information Matrices}\label{sec:two_MI}

\subsubsection{Cross-Mutual Information Matrix}
We first analyze the learning dynamics by designing a cross-mutual information matrix between a network under training (denoted by $\mathcal{S}$) and a reference network (denoted by $\mathcal{T}$). Here, the training network refers to the network during the training whose parameters are updated in each epoch, while the reference network refers to the final learned network at the end of training in which the parameters are fixed. We consider the entire network $F$ as the composition of $L$ layers, where $F=f_{1} \circ \ldots \circ f_{L}$. Given an input $X \in \mathbb{R}^{n\times d}$ with $n$ samples and $d$-dimensional features, the intermediate representation in the $i$-th layer can be denoted as $\bm{z}_i \in \mathbb{R}^{n\times d_i}$ with $d_i$ dimension, where $\bm{z}_i=f_i(\bm{z_{i-1}})$. 

\begin{figure*}[h!]
\centering
\includegraphics[width=0.8\textwidth]{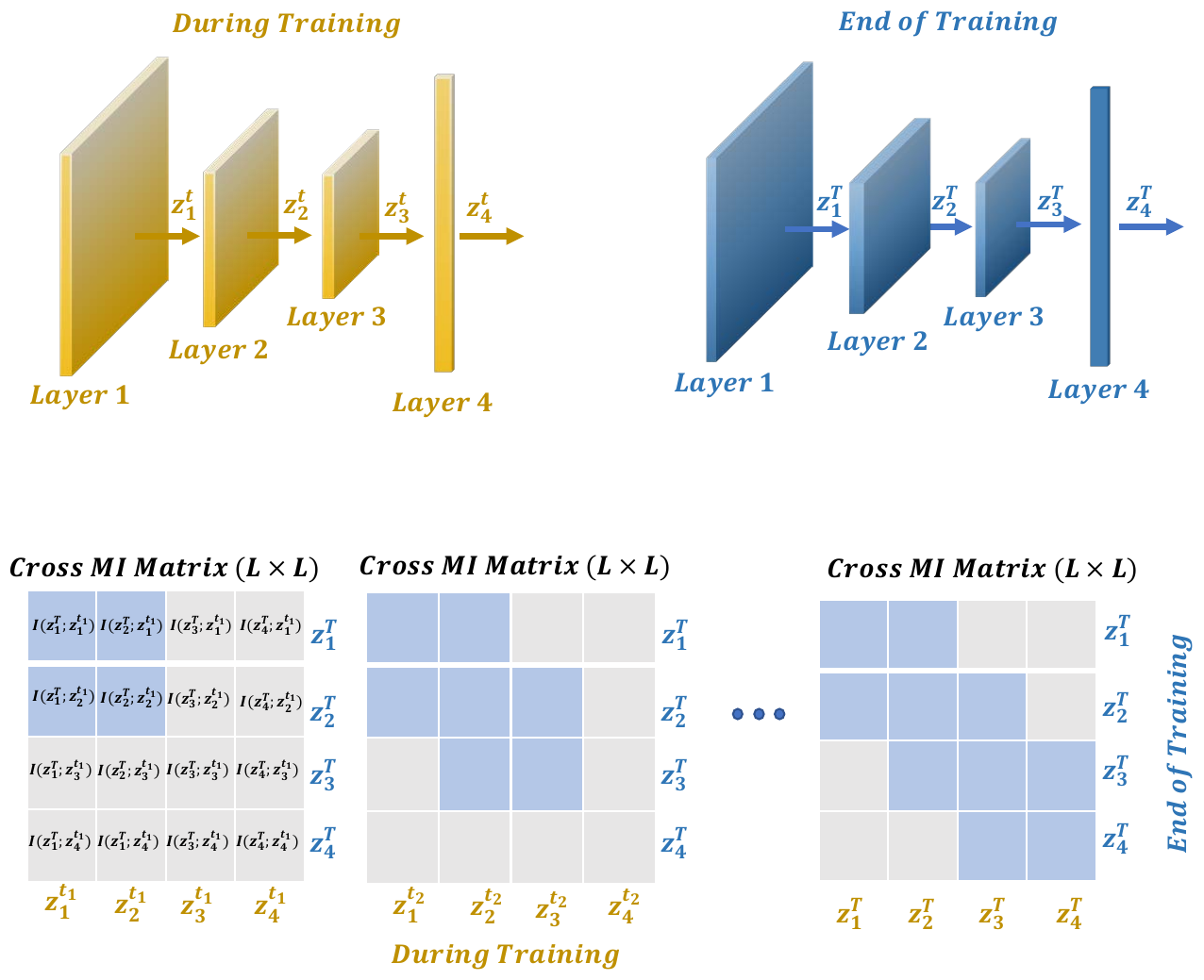}
\caption{Illustrating E2EBP learning dynamic based on the cross MI matrix. We assume both $\mathcal{S}$ and $\mathcal{T}$ have $4$ layers for simplicity.}
\label{fig:cross_MI_matrix}
\end{figure*}

We compute the mutual information value between the $i$-th layer representation in $\mathcal{S}$ at training epoch $t$ (denoted by $\bm{z}_i^t$) and the $j$-th layer in $\mathcal{T}$ (denoted by $\bm{z}_j^\mathcal{T}$). 
This procedure results in a cross-mutual information matrix $G^{t}$, the size of which is $L\times L$ and the $(i,j)$-th entry is $G^{t}_{i,j} = I(\bm{z}_i^{t};\bm{z}_j^{\mathcal{T}})$.
Intuitively, \( G^t \) quantifies the shared information between all layer representations at epoch \( t \) and those at the end of training. In other words, \( G^t \) measures how similar or dependent the network is after \( t \)-epochs of training compared to the fully trained network.
A high value in $G^t$ implies that the network after $t$-epoch of training is highly similar or relevant to the final trained network; whereas a low value indicates dissimilarity.
For clarity, we illustrate this procedure in Figure~\ref{fig:cross_MI_matrix}, where we assume that both \( \mathcal{S} \) (top left figure) and \( \mathcal{T} \) (top right figure) consist of 4 layers. The cross-MI matrices are evaluated at training epochs \( t_1 \), \( t_2 \), and at the end of the training.


\begin{figure}[h!]
	\setlength{\abovecaptionskip}{0pt}
	\setlength{\belowcaptionskip}{0pt}
	\centering
	
	\subfigure[VGG16]{
		\includegraphics[width=10cm]{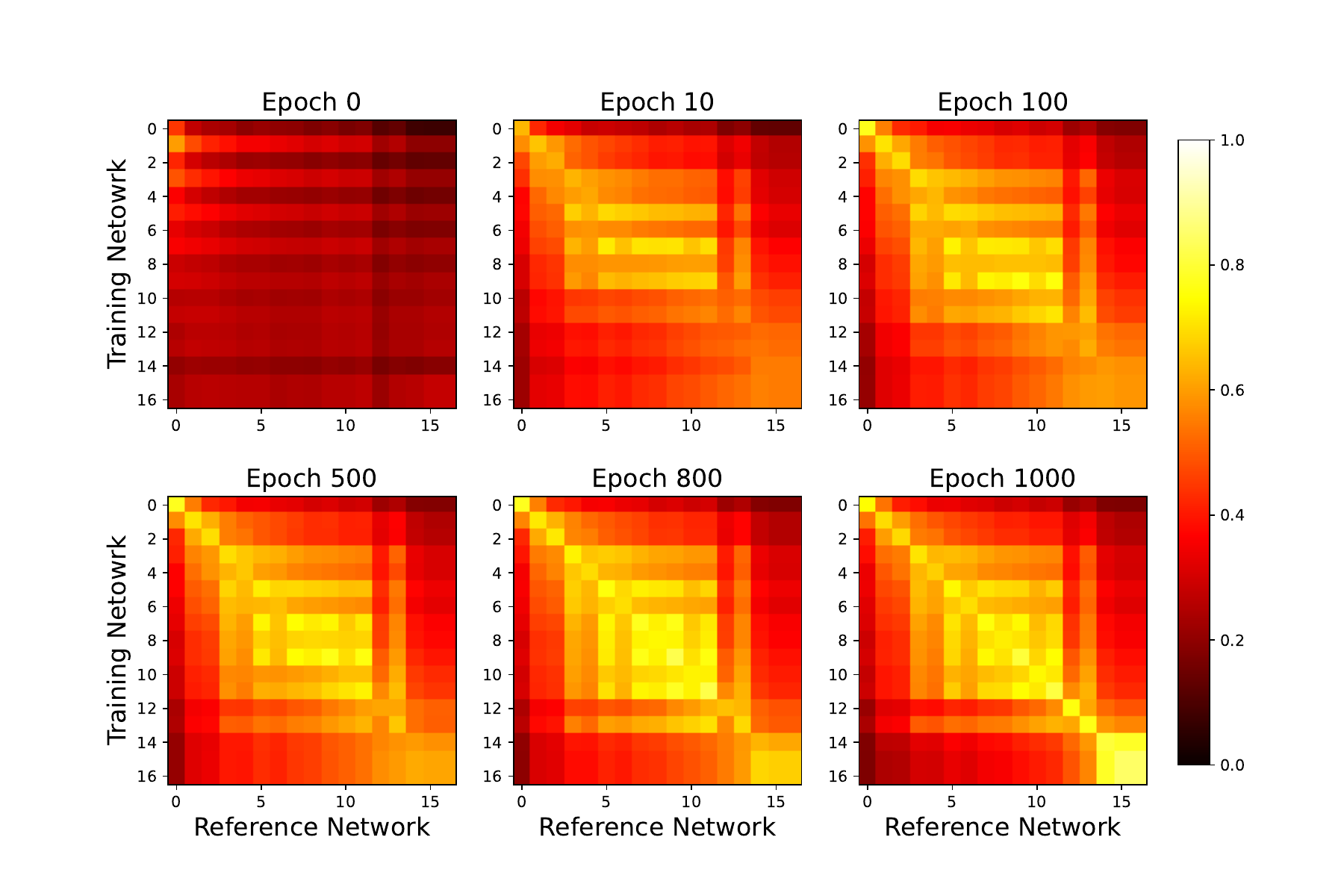}}
	\subfigure[ResNet18]{
		\includegraphics[width=10cm]{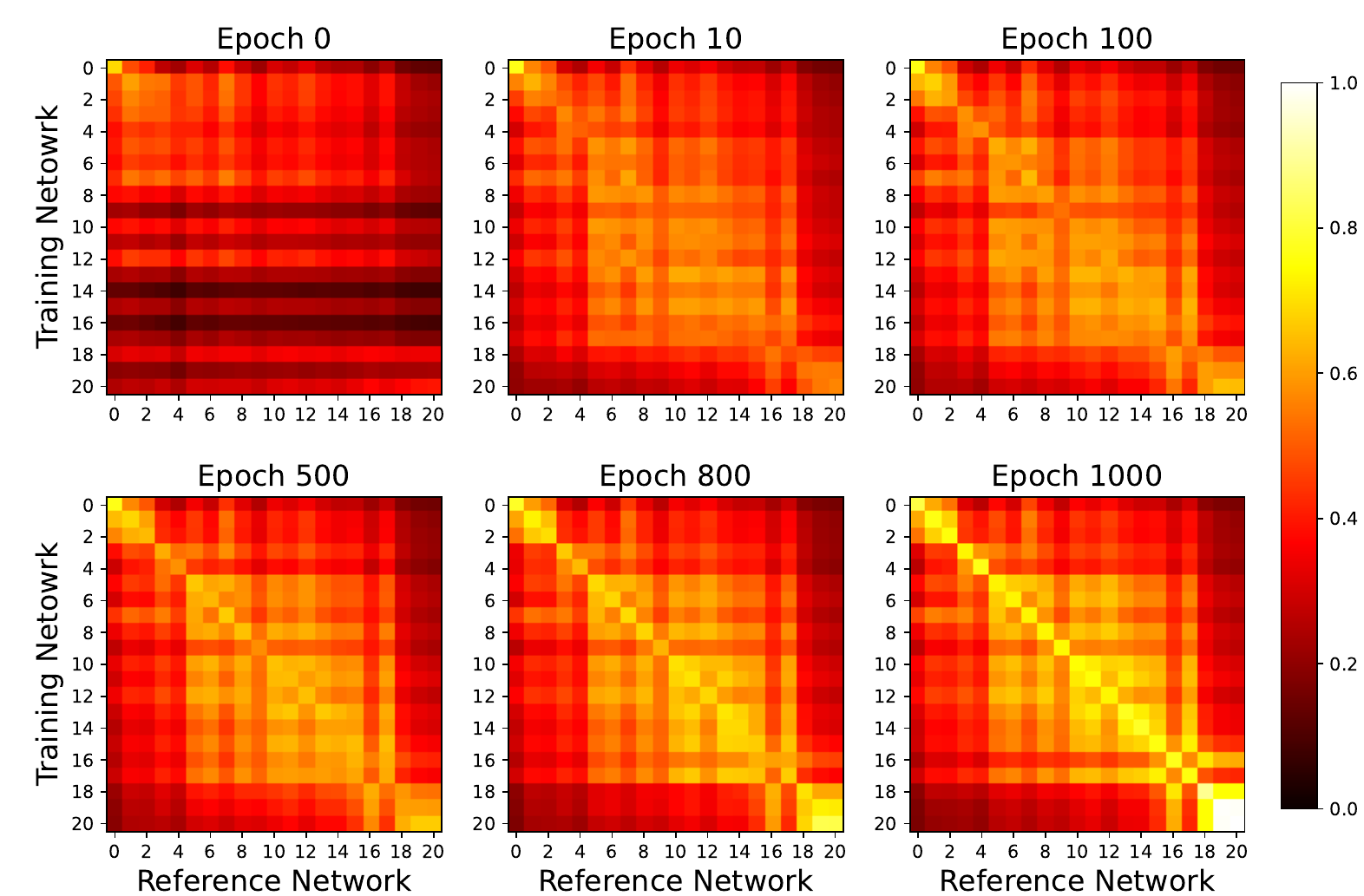}}
	\caption {Cross-mutual information matrix of VGG16 and ResNet18 trained on CIFAR-10. Each pane is a $L\times L$ matrix, the x-axis denotes all layers' output for reference network and y-axis represents all layers' output for training network, and each entry of the matrix represents the mutual information between two layers. The larger value means high dependency.}
\label{dynamic_training}
\end{figure}

Throughout this paper, we measure mutual information values with the matrix-based Matrix-based R\'enyi’s $\alpha$-order entropy functional~\cite{giraldo2014measures,yu2019multivariate} (see Appendix~\ref{MI_estimator} for more details), which is computationally efficient and scalable to high-dimensional variables.

The cross-mutual information matrices of VGG-16 and ResNet18 trained on CIFAR-10 for a total of $1,000$ epochs with SGD are shown in Figure \ref{dynamic_training}. As can be seen, the values of the diagonal elements in shallow layers rapidly converge to large values within a short period of time, while the diagonal elements in deeper layers increase more gradually over a longer training period. These figures suggest that the end-to-end training of deep neural networks progresses sequentially, starting from the shallow layers and moving toward the deep layers. Our observations are consistent with~\cite{raghu2017svcca}, which relies on Canonical Correlation Analysis (CCA), a method that quantifies only linear dependence.

\subsubsection{Auto-Mutual Information Matrix}
We propose an alternative method to analyze the learning dynamics of E2EBP by visualizing the trajectory of different layers' outputs in a 2-D plane. Unlike the cross-mutual information matrix, we directly construct a pairwise distance matrix \( D^t \), with size \( L \times L \), using an auto-mutual information matrix. The \((i,j)\)-th entry of this matrix is given by \( I(z_i^t, z_j^t) \) rather than \( I(z_i^t, z_j^\mathcal{T}) \). Intuitively, the auto-mutual information matrix captures all pairwise dependencies between the layer representations at training epoch \( t \).

\begin{figure}[h!]\label{MI_matrix}
\centering
\includegraphics[width=1\textwidth]{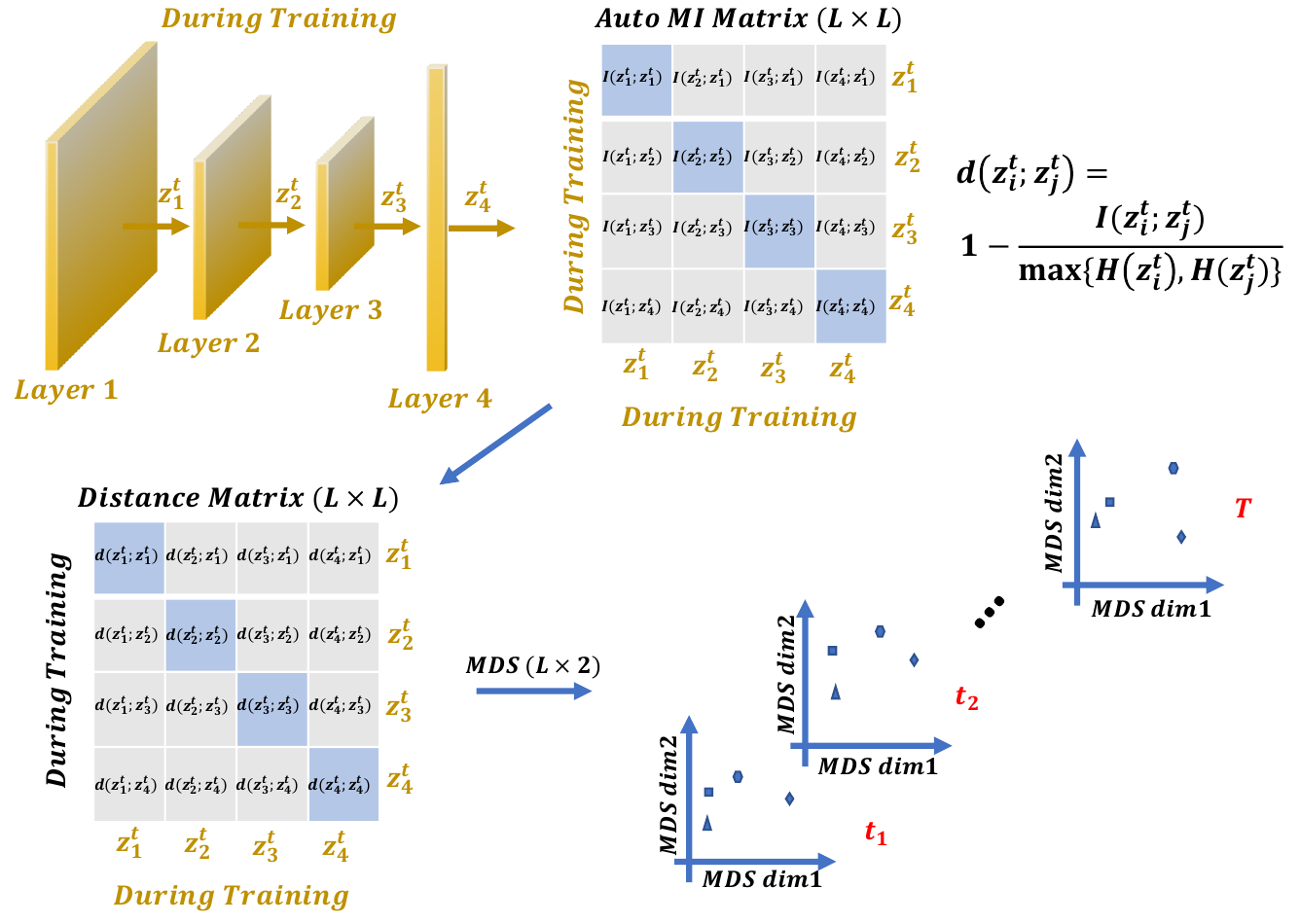}
\caption{Illustrating layer's trajectory of E2EBP during training with MDS based on the auto-MI matrix.}
\label{fig:MI_matrix}
\end{figure}

The distance between two layers' output at the $t$-th epoch $d(\bm{z}_i^{t};\bm{z}_j^{t})$ is defined as:
\begin{equation}\label{normalzie_MI}
D^t_{i,j} =  1 -\frac{I(\bm{z}_i^{t};\bm{z}_j^{t})}{\max\{H(\bm{z}_i^{t}),H(\bm{z}_j^{t})\}},
\end{equation}
where \( H \) denotes the entropy, and the second term is the normalized mutual information, which ranges from 0 to 1. \( D^t_{i,j} \) is a distance metric, as demonstrated in~\cite{6313441, chen2009similarity}. We then project the pairwise distance matrix \( D^t \) into a 2-D plane using multidimensional scaling (MDS). The coordinates of different layers reflect their pairwise distances. This allows us to continuously observe the trajectories of different layers throughout the entire training process. An illustrative figure of this procedure is demonstrated in Figure~\ref{fig:MI_matrix}.

\begin{figure}[htbp]
	\setlength{\abovecaptionskip}{0pt}
	\setlength{\belowcaptionskip}{0pt}
	\centering
	
	\subfigure[]{
		\includegraphics[width=8cm]{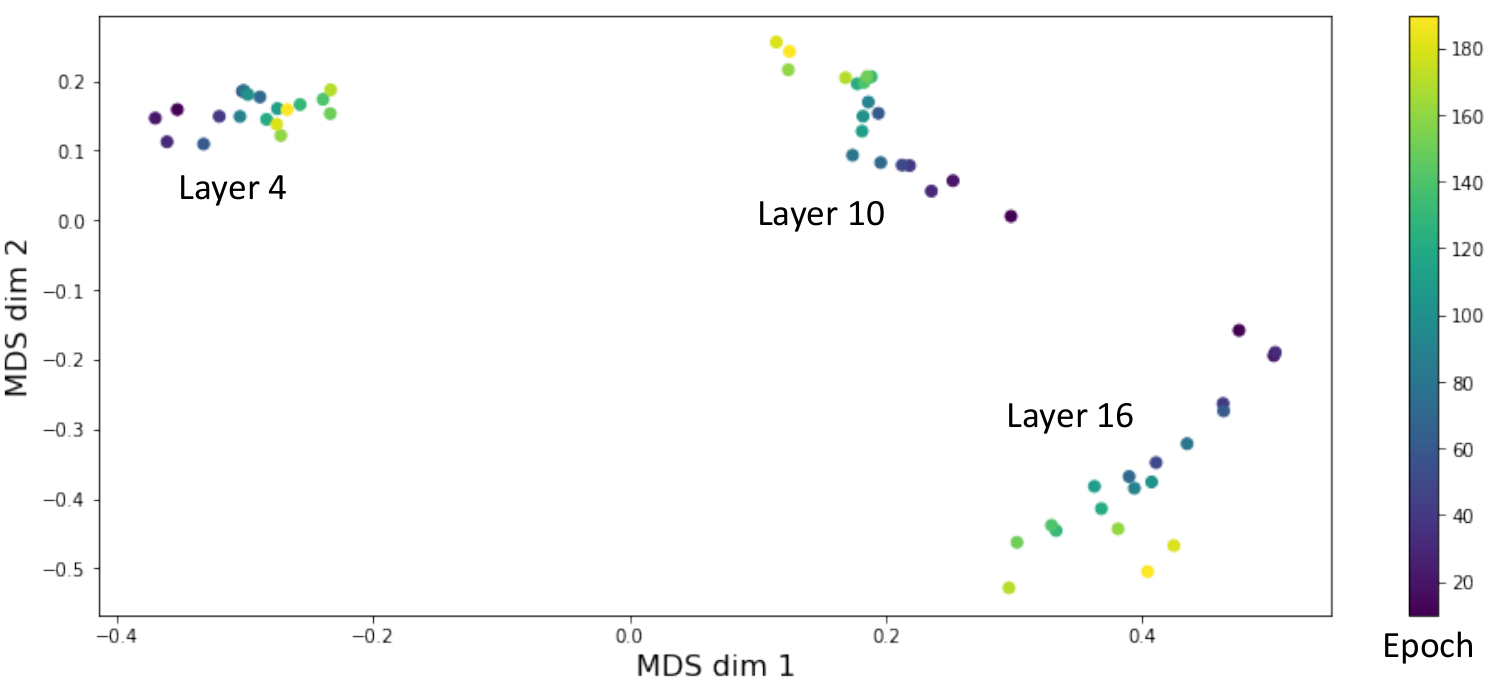}}
	\subfigure[]{
		\includegraphics[width=8cm]{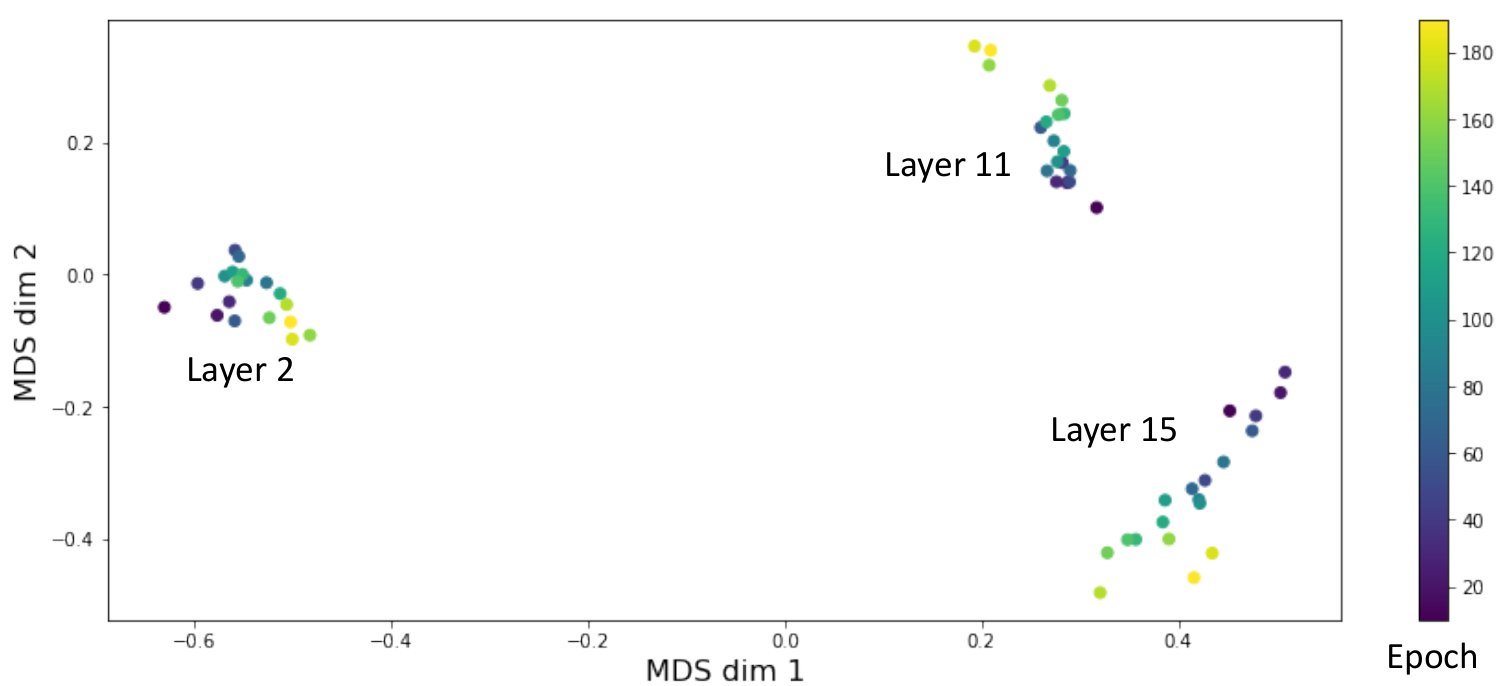}}
	\caption{VGG16 learning dynamic with MDS measurement for shallow, middle and deep layer. (a) Shallow 4 layer, middle 10 layer and deep 16 layer. (b) Shallow 2 layer, middle 11 layer and deep 15 layer.}
	\label{MDS}
\end{figure}

We plot trajectories of six layers from shallow ($2$-nd and $4$-th), middle ($10$-th and $11$-th), to deep ($15$-th and $16$-th) layer of VGG-16 architecture during the end-to-end training in Figure~\ref{MDS}. We observe that the coordinates of the shallow and middle layers do not change significantly during training, whereas the coordinates of the deeper layers show substantial movement. This observation indicates that the shallow and middle layers are trained much more quickly than the deeper layers, which further supports our claim that, even when a network is trained with E2EBP, its internal learning dynamics still follow a layer-wise convergence pattern.

\subsection{Modified Information plane}\label{sec:IP}
In this section, we use the information plane to visualize \( I(z_{l-1};z_{l}) \) (the mutual information between the output of the \( l \)-th layer, \( z_{l} \), and the output of the previous layer, \( z_{l-1} \)) and \( I(z_{l};y) \) (the mutual information between the current layer's output, \( z_{l} \), and the target \( y \)) for VGG16 and ResNet10 trained on CIFAR-10 across 1,000 training epochs.

\begin{figure}[h!]
	\centering
	\subfigure[VGG16]{
		\includegraphics[width=13cm]{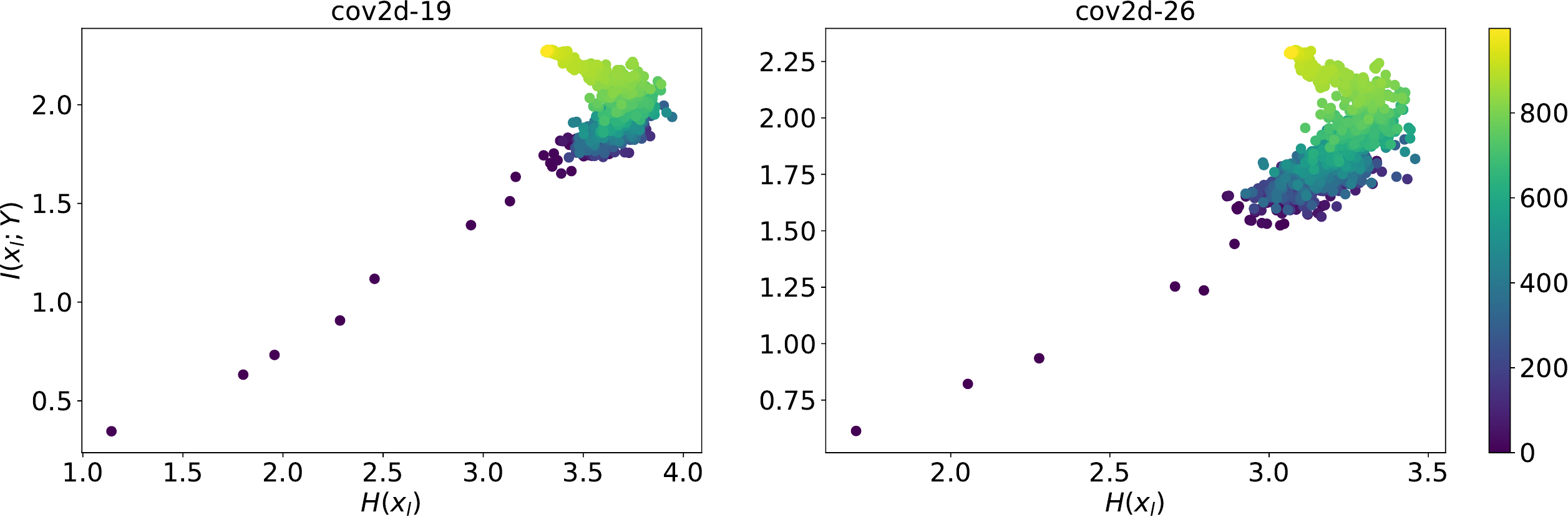}}
        \\
	\subfigure[ResNet18]{
		\includegraphics[width=13cm]{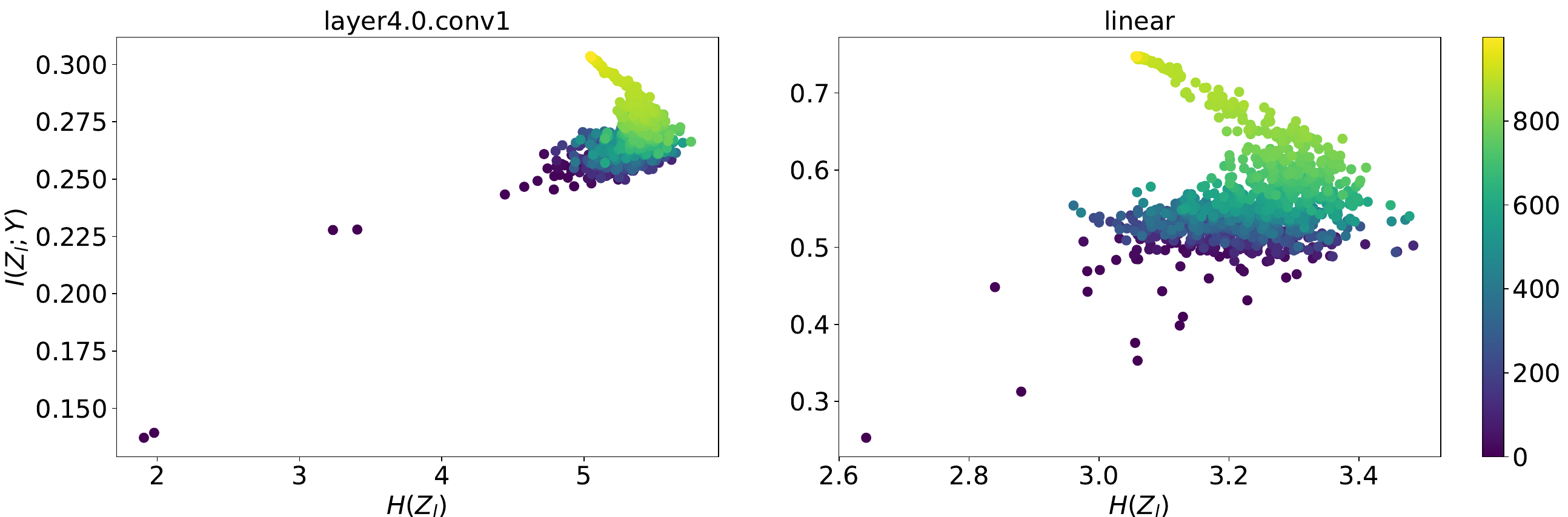}}
	\caption {Information plane of VGG16 and ReNet18 on CIFAR-10 with different convolution and linear layers (\emph{i.e.,} shallow layer: "cov2d-19" "layer4.0.conv1", deep layer: "cov2d-26", "linear"). Different colors correspond to the number of training epochs.}
\label{IB_observe}
\end{figure}

For a deterministic neural network such as VGG16, we can replace $I(z_{l-1};z_{l})$ with $H(z_{l})$\footnote{
$I(z_{l-1};z_{l}) = H(z_{l}) - H(z_{l}|z_{l-1})$ and $H(z_{l}|z_{l-1})=0$ because there is no uncertainty in the mapping from $(l-1)$-th layer to the $l$-th layer~\cite{strouse2017deterministic}.
} and plot $H(z_{l})$ with respect to $I(z_{l};y)$ in Figure~\ref{IB_observe}. We observe that $I(z_{l};y)$ continues to increase during the whole training epochs while $I(z_{l-1};z_{l})$ increases at the beginning and followed by an obvious decrease until convergence. The increase of $I(z_{l};y)$ is expected from the cross-entropy loss minimization. The trajectory of $H(z_{l})$ indicates the fitting (information increasing) and compression phase (information decreasing) during the E2EBP. 
Hence, when designing a local learning objective for each layer, the objective should include an information compression term to explicitly model this phenomenon.

\textbf{Summarization of empirical observations on E2EBP}: The experiments above reveal that DNNs converge sequentially from the shallower layers to the deeper layers, even when trained using E2EBP. Moreover, the training of each layer within DNNs under the traditional E2EBP adheres to the IB principle, where the hidden layer representations first increase and then decrease, effectively removing redundant or noisy information from the input \( X \) that does not contribute to classifying the label \( Y \). Based on these observations, we consider whether it is possible to explicitly train a DNN from shallow layers to deeper layers in a manner that follows exactly the IB principle. We shall give an answer in the next section.

\section{Layer-wise Training with Deterministic Information Bottleneck}\label{section4}
We utilize the standard convolutional and fully connected model structures, but each layer is trained by a local learning signal motivated by deterministic information bottleneck, instead of globally back-propagating errors. According to Figure \ref{IB_observe}, we observe that there exists a compression phase during which $H(z_{l})$ decreases, while the $I(z_{l};y)$ continue increases until the end of training. This observation motivates us to design a new layer-wise training method such that the mutual information $I(z_{l};y)$ is maximally preserved whereas the redundant information in $z_{l}$ is dropped out by minimizing $H(z_{l})$.

\subsection{Deterministic Information Bottleneck (DIB)}\label{DIB_define}

Let $X$ be an input signal, $Z$ represents the latent representation of $X$, and $Y$ refers to the corresponding label. Given the joint distribution $p(x,y)$, the original IB objective~\cite{tishby2000information} seeks to find an encoding distribution $q(z|x)$ through the following optimization problem:
\begin{equation}\label{IB_lagrangian}
\operatorname*{min}_{q(z|x)} \mathcal{L}_{\textmd{IB}}[Z]=-I(Z;Y)+\beta I(Z;X),
\end{equation}
in which $I(Z;Y)$ characterizes the \emph{sufficiency} of $Z$ to address a given task, whereas $I(Z;X)$ characterizes the \emph{minimality} of $Z$ with respect to $X$. $\beta$ is a hyperparameter that balances the trade-off between $I(Z;Y)$ and $I(Z;X)$. In this sense, IB can be interpreted to learn a minimum sufficient representation.

For a deterministic neural network, we could replace the mutual information $I(Z;X)$ with entropy $H(Z)$. This is because $I(Z;X) = H(Z)-H(Z|X)$ and $H(Z|X)=0$~\cite{kirsch2020unpacking,ahuja2021invariance}. Therefore, the IB objective naturally reduces to the deterministic IB objective~\cite{strouse2017deterministic}: 
\begin{equation}\label{DIB_lagrangian}
\operatorname*{min}_{q(z|x)} \mathcal{L}_{\textmd{DIB}}[Z]=-I(Z;Y)+\beta H(Z).
\end{equation}

Additionally, we have \(-I(Z;Y) = -H(Y) + H(Y|Z)\). Therefore, minimizing \(-I(Z;Y)\) is equivalent to minimizing \(H(Y|Z)\), as the entropy \(H(Y)\) of the ground truth labels \(Y\) is a constant and does not depend on the network parameters.

Given the training set \( \{ x_{i}, y_{i} \}_{i=1}^{N} \), let \( p_\theta(z|x) \) and \( p_\theta(y|z) \) denote the unknown distributions we aim to estimate, parameterized by \( \theta \). The term \( H(Y|Z) \) can be further approximated as the empirical cross-entropy (CE) as shown in Eq. (\ref{H_YT}):
\begin{equation}\label{H_YT}
    H(Y|Z)\simeq \mathbb{E}_{x,y\sim p(x,y)}\left[\mathbb{E}_{z\sim p_\theta(z|x)}\left[-\log p_\theta(y|z)\right]\right] = \frac{1}{N}\sum_{i=1}^N \mathbb{E}_{z\sim p(z|x_i)}\left[-\log p(y_i|z)\right],
\end{equation}

Therefore, the DIB objective reduces to a basic cross-entropy loss, regularized by a weighted differentiable entropy term \( H(Z) \), as defined in Eq.~(\ref{DIB}):
\begin{equation}\label{DIB}
\mathcal{L}_{\text{DIB}}=\text{CE}(\hat{Y};Y)+\beta H(Z).
\end{equation}

\subsection{Greedy layer-wise training with DIB}
\subsubsection{Architecture Formulation}
\begin{figure*}
\centering
\includegraphics[width=1\textwidth]{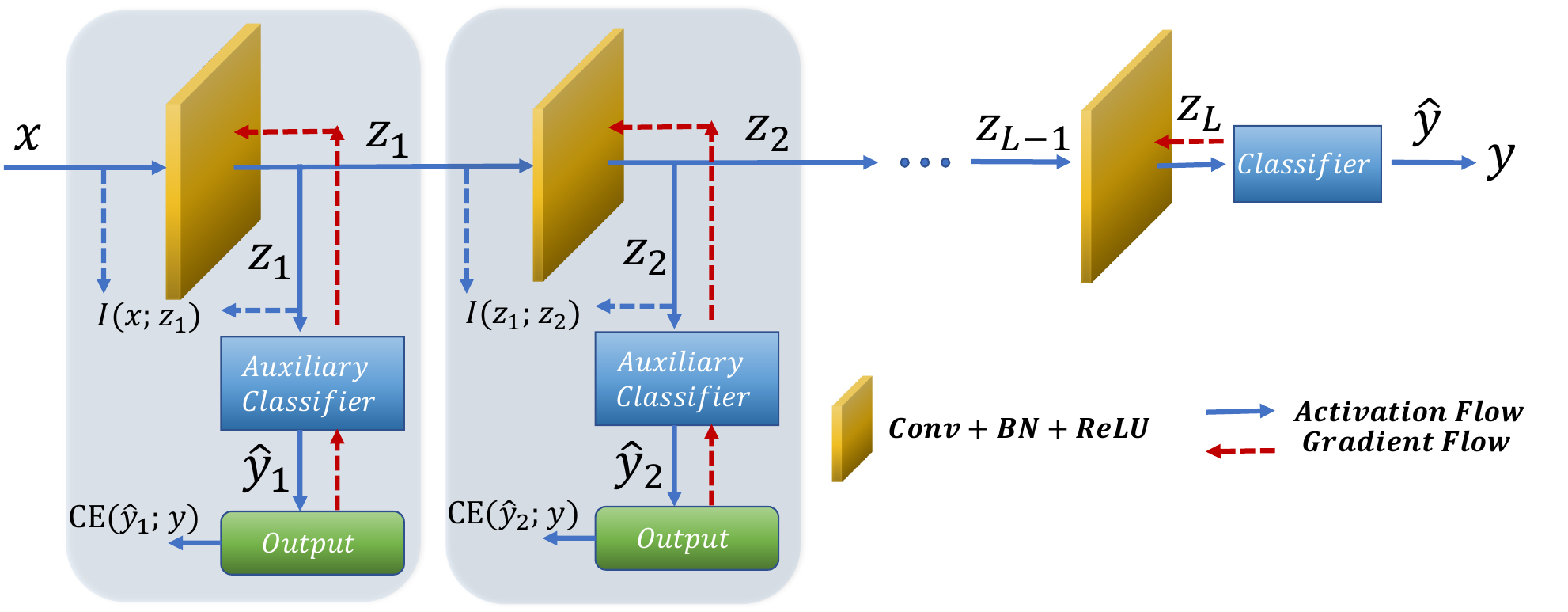}
\caption{Diagram of our proposed method. Each layer is jointly
trained with the auxiliary classifier with cross-entropy loss related to the label, and regularized with the entropy of the layer's output.}\label{pipeline}
\end{figure*}

Our architecture has $L$ blocks (see Figure \ref{pipeline}), which are trained in succession.  For an input signal $x$, it is propagated from the first hidden layer to the $l$-th hidden layer, generating $z_l$. Each $z_l$ feeds into an auxiliary classifier to obtain $\hat{y}_l$, the intermediate classification output solely based on $z_l$. At the \(l\)-th block, the network component is denoted as \(f_{\theta_l}\), parameterized by \(\theta_l\), and consists of standard convolutional layers followed by a batch normalization layer and a ReLU activation function. The corresponding auxiliary classifier for the \(l\)-th block is represented as \(g_{\phi_l}\), parameterized by \(\phi_l\), which can be a simple fully connected layer or a convolutional layer. Formally, given an input signal $x$, we iterate as follows:
\begin{equation}
\left\{\begin{aligned}
z_{l} &=f_{\theta_l}(z_{l-1}), \ l= 1,2,...,L\\
\hat{y}_{l} &=g_{\phi_l}(z_{l}) \in \mathbb{R}^c,
\end{aligned}\right.
\end{equation}
where $c$ is the number of classes.

\subsubsection{Training with Deterministic Information Bottleneck} 
Our training procedure is layer-wise. Specifically, to train the $l$-th layer, we only update layer parameters $\theta_l$ and corresponding auxiliary classifier parameters $\phi_l$ by a DIB objective, whereas all other parameters are fixed. 
In particular, the learning objective for the $l$-th hidden layer is represented as:
\begin{equation}\label{DIB_loss}
\begin{split}
\mathcal{\hat{R}}_{l}(z_l;\theta_l,\phi_l)&=\beta I\big(z_{l-1};z_{l}\big)-I(z_{l};y)=\beta H(z_{l})-I(z_{l};y)\\
&=\beta\underbrace{\frac{1}{1-\alpha}\log_{2}\left(\sum_{i=1}^{N}\lambda _{i}(A)^{\alpha}\right)}_{\textcircled{\small{1}}}+\underbrace{\frac{1}{N}\sum_{i=1}^N \mathbb{E}_{\mathbf{z}\sim p(\mathbf{z}|\mathbf{x}_l)}\left[-\log p(y_i|z_l^i)\right]}_{\textcircled{\small{2}}}.
\end{split}
\end{equation}

The first term in Eq.~(\ref{DIB_loss}) is estimated with the R\'enyi's $\alpha$-order entropy functional (see part \textcircled{\small{1}}). Specifically, given a minibatch of samples $\{x_i,y_i\}_{i=1}^{B}$ of size $B$, we obtain the hidden layer representations $\{(z_l)_i\}_{i=1}^B$ at the $l$-th layer and construct a kernel Gram matrix $K\in \mathbb{R}^{B\times B}$ with Gaussian kernel of kernel width $\sigma$ as $K_{i,j} = \kappa\left((z_l)_i,(z_l)_j\right)$ and $\kappa = \exp(-\frac{\| (z_l)_i-(z_l)_j \|_2^2}{2\sigma^2})$.
$A=K/\tr{K}$ is the trace normalized kernel Gram matrix and $\lambda _{i}(A)$ denotes the $i$-th eigenvalue of $A$. 
The second term $I(z_l;y)$ in Eq.~(\ref{DIB_loss}) is approximated with the cross-entropy as discussed earlier (see parts \textcircled{\small{2}}). The full algorithm of our layer-wise training is presented in Algorithm~\ref{alg:cnn}.



\begin{algorithm}
\caption{Layer Wise CNN}\label{alg:cnn}
\begin{algorithmic}
\Require Training samples $\{ x_i, y_i \}_{i=1}^N$
\Ensure $\{\theta_l\}_{l=1}^L$ and $\phi_L$.
\While{stopping criterion not met}
\State Sample a minibatch of $B$ samples from the training set.
\For{$j \in 1,\cdots, L$}
    \State $(\theta_l^*,\phi_l^*) = \argmin_{\theta_l,\phi_l} \mathcal{\hat{R}}_{l}(z_l;\theta_l,\phi_l) $
\EndFor
\EndWhile
\end{algorithmic}
\end{algorithm}

The most relevant work is the HSIC bottleneck, which also employs the IB principle to train deep classification networks without error backpropagation. In addition to differences in the mutual information estimator and the IB objective, we would like to further clarify that the training method and network architecture are also distinct. Specifically, our method utilizes greedy layer-wise training, while the HSIC bottleneck adopts multi-round layer-wise training, which is more computationally complex and less memory efficient. Regarding the network architecture, HSIC bottleneck does not attach an auxiliary model to each layer’s output, whereas our method leverages an auxiliary model to map the higher-dimensional latent representations to a lower dimension, making it more convenient to design and compute the proxy objective function.

\section{Experiments}\label{sec:experiments}
\subsection{Implementation details}
To validate the scalability of our proposed method, we evaluate it in both simple and complex architectures. For a simple CNNs experiment, we trained a model with five convolutional layers (256 filters with $3\times 3$ kernel size) without any down-sampling layer. We selected the auxiliary classifier as only one fully connected layer, the size of which is 1024. 
Each convolutional block consists one convolutional layer, one batch-normalization layer, and ReLU activation layer. We train each convolutional block with an auxiliary classifier attached by an averaging pooling layer. We optimize each layer with SGD using a momentum of 0.9, weight decay of $5e-4$, a batch-size of 256. A learning rate of 0.1 is used initially, followed decreasing by a factor of 0.2 every 15 epochs for a total of 100 epochs in each layer. For more complex architectures (i.e., VGG and ResNet), we set the auxiliary classifier model as two convolutional and one fully connected layer. Each layer is optimized with SGD using a momentum of 0.9, weight decay of $1e-4$. We set batch-size as 128 and each layer starts with 0.1 learning rate and then decays by 0.1 every 20 epochs, for a total of 50 epochs. We use the standard data augmentation, flipping, and cropping for CIFAR-10 and CIFAR-100 datasets. 

\subsection{Performance Comparison and Analysis}
\subsubsection{Simple CNNs on CIFAR-10}

\begin{figure}[htbp]
	\setlength{\abovecaptionskip}{0pt}
	\setlength{\belowcaptionskip}{0pt}
	\centering
	
	\subfigure[]{
		\includegraphics[width=5cm]{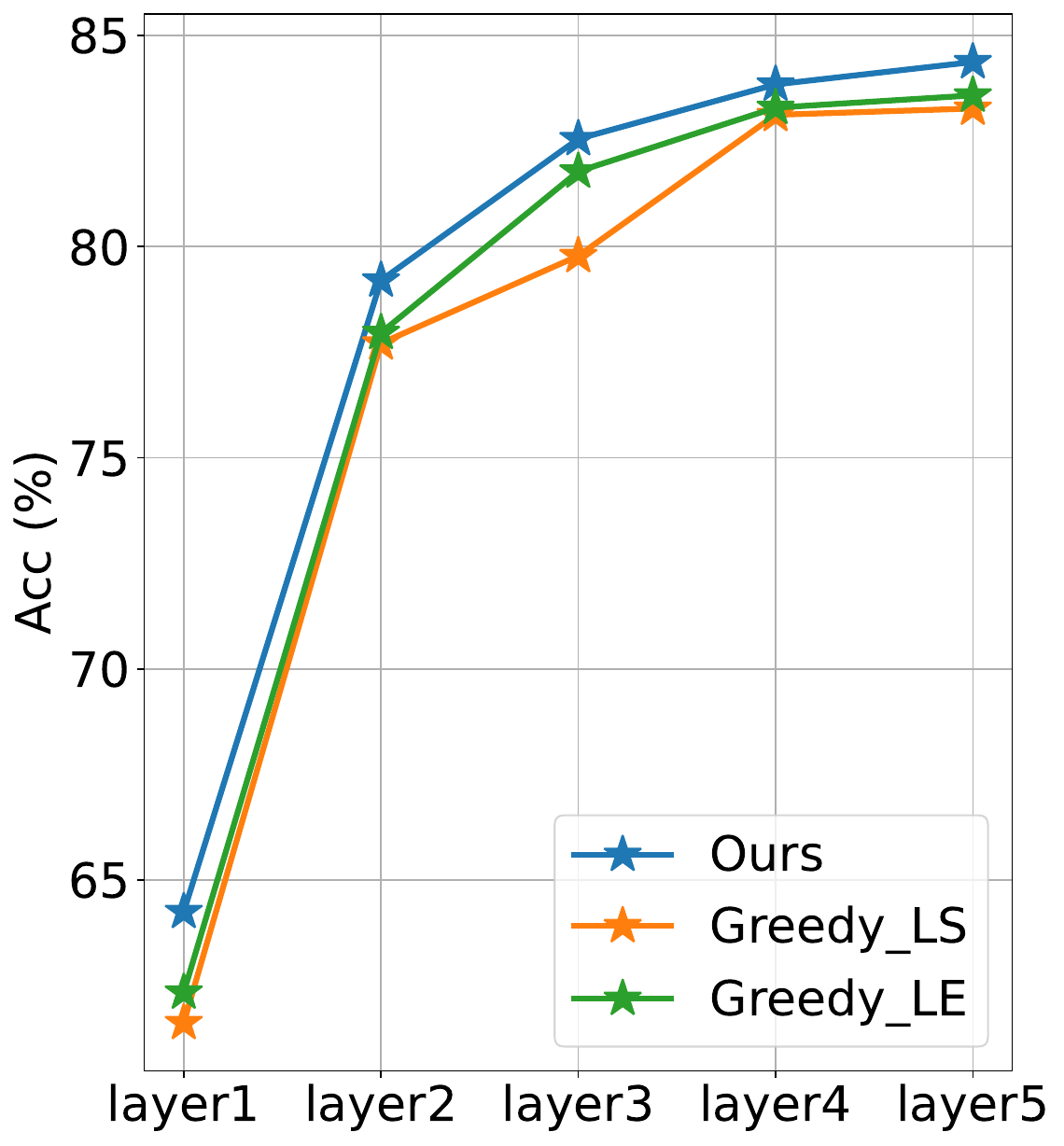}}
	\subfigure[]{
		\includegraphics[width=5cm]{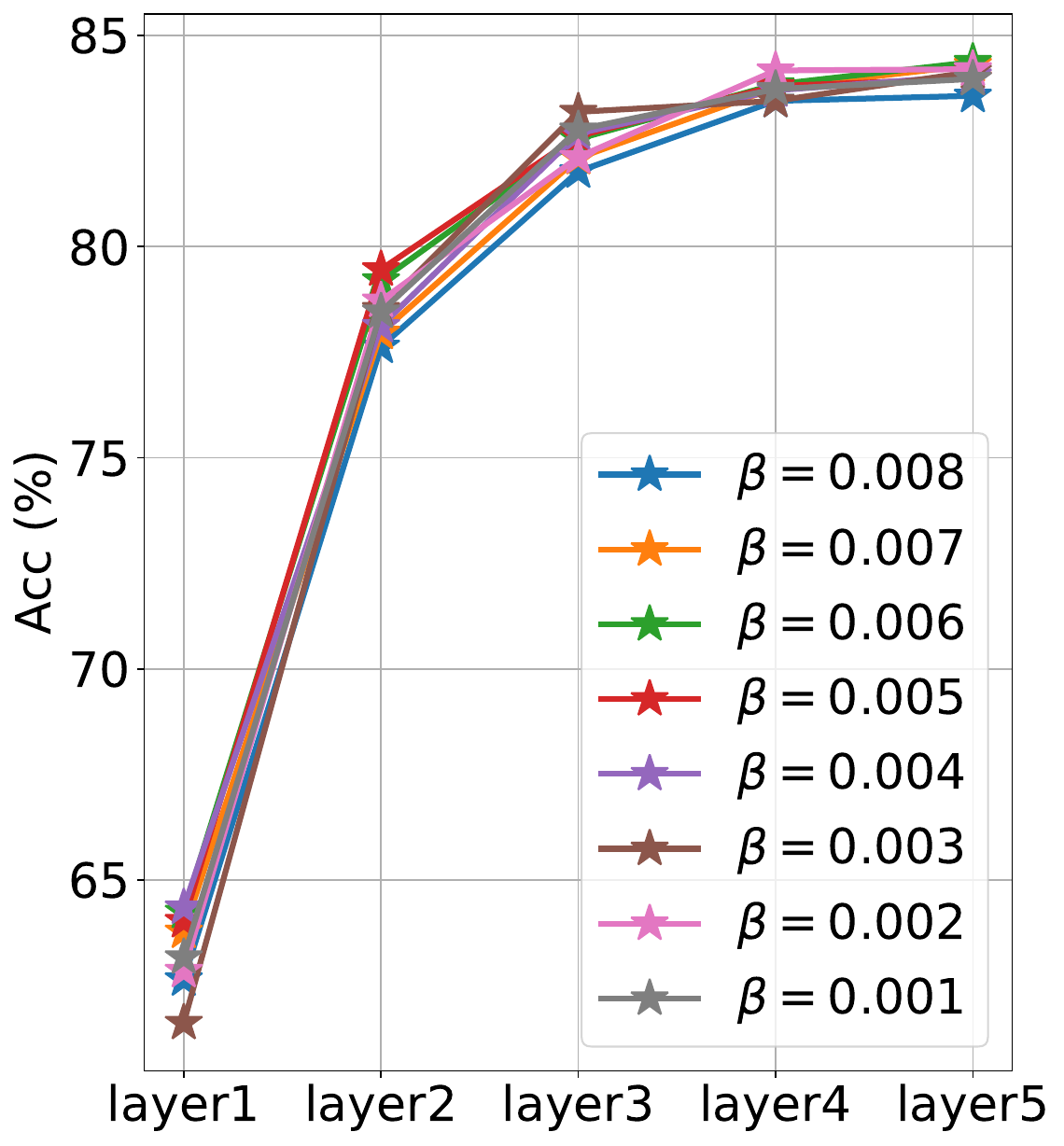}}
	\caption {\small (a) Accuracy with different layers on CIFAR-10 test set. (b) Proposed method results with different $\beta$.}
	\label{fig2}
\end{figure}

In this section, we evaluated our proposed and other baseline methods with simple 5-layers CNN on the CIFAR-10 dataset.  Specifically, according to their original setting, we set $\beta=0.01$ and $\beta=10$ for Greedy LE and HSIC bottleneck methods.  To avoid hyperparameter tuning, we use a fixed value of $\beta$ for different layers. We find that the performance is stable when $\beta$ located between 0.001 to 0.01, thus we sweep the $\beta$ in this range and found that $\beta=0.006$ provides the best performance shown in Fig.~\ref{fig2}(b). 

\begin{table}[ht]
\centering
\caption{Test accuracy (\%) on CIFAR-10 with 5-layer CNN. The best performance is in bold and the second best performance is underlined.}
\begin{tabular}[t]{lc}
\hline
\bf{Method}&\bf{Test Acc} (\%)\\
\midrule
Backprop (SGD) &\underline{84.63}\\
\midrule
Greedy LS &83.27\\
Greedy LE&83.58\\
HSIC bottleneck&61.56\\
NIK&60.72\\
\midrule
{\bf Ours}&\textbf{84.66}\\
\hline
\end{tabular}
\label{simple_CNNs}
\end{table}%

\begin{figure*}
\centering
\includegraphics[width=1\textwidth]{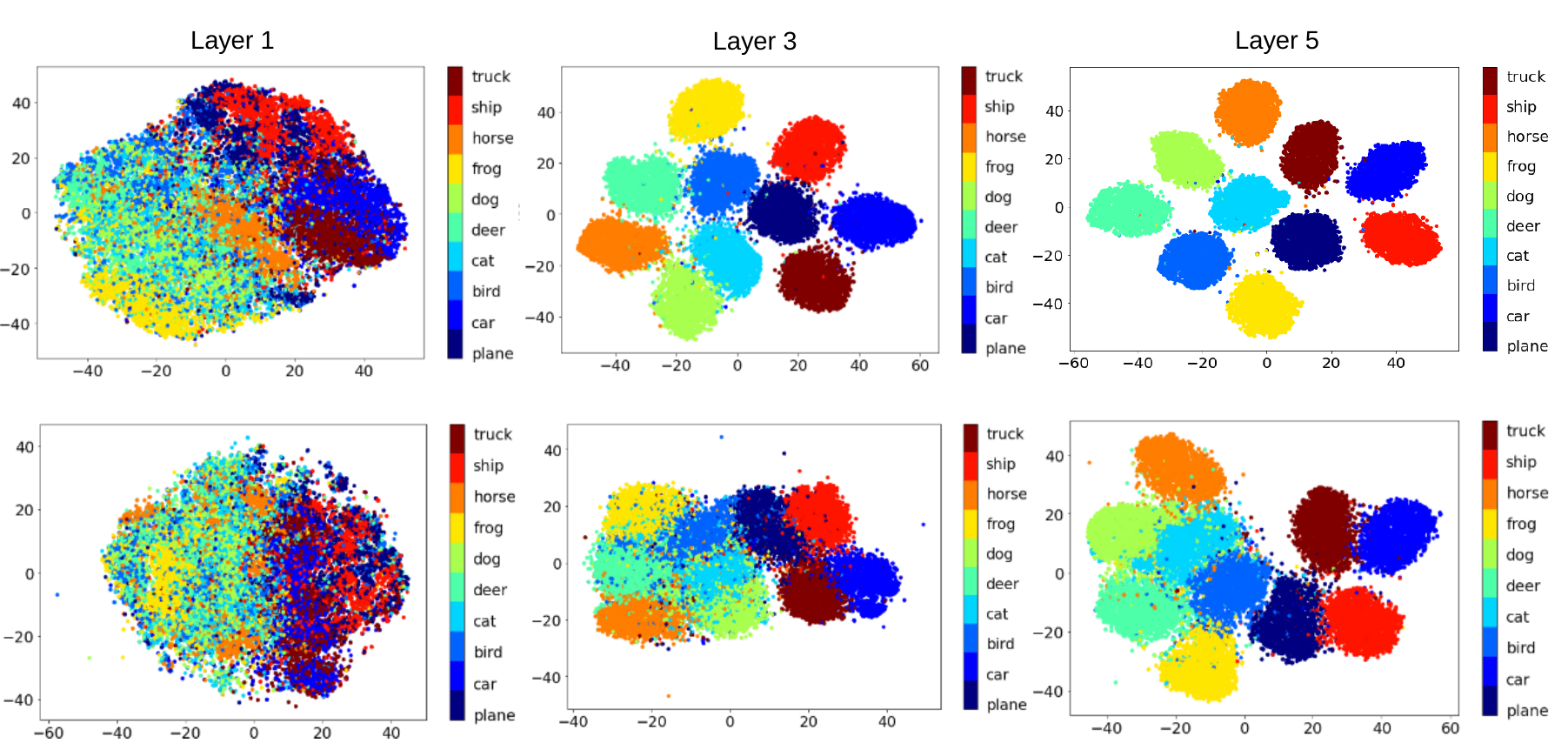}
\caption{\small The latent representation for different layer's output on CIFAR10 validation set in 5-layer CNN. The first row represents the results for greedy layer-wise training with DIB. The second row represents results for training with only cross-entropy loss (i.e., without DIB).}\label{fig3}
\end{figure*}

We report the results trained with 5-layer CNN in Table~\ref{simple_CNNs} and different number of layers in Figure~\ref{fig2}(a). Our proposed method outperforms existing proxy objective based layer-wise training methods and is the only one that is comparable to the performance of end-to-end training method (with a difference on accuracy less than $1\%$. Greedy LS and Greedy LE approaches are auxiliary-based methods that achieve better performance than NIK and HSIC bottleneck. This indicates that an explicit auxiliary classifier is more effective than kernel alignment mechanisms to transfer the label information to guide the training of intermediate layers. Our proposed method achieves better results than Greedy LS and Greedy LE because we not only incorporate the label information for training each layer but also leverage the information bottleneck principle to discard the redundancy within the layers. In addition, we also illustrate the latent representation of the CIFAR-10 validation set in Figure \ref{fig3} to show the effectiveness of the DIB principle. We observe that the latent representations produced by our proposed method are more separable compared to those generated without DIB.

\subsubsection{Large-scale Experiments}

 We further evaluated our method with more complex architectures with the CIFAR-10 and CIFAR-100 datasets. The VGG-11 architecture contains 8 convolutional layers, 1 average-pooling layer, and 2 fully connected layers, while the VGG-16 architecture contains 13 convolutional layers, 1 average-pooling layer, and 2 fully connected layers. Thus, we train total 8 layer blocks for VGG-11 and total 13 layer blocks for VGG-16. We consider each layer block as one convolutional layer, batch-normalization, and ReLU activation layer. For ResNet-18 network, we consider each layer block as two convolutional layers with batch normalization and ReLU activation, and a skip connection and train total 8 layer blocks. Similar to the previous setting, each layer block is trained jointly with the auxiliary classifier. We summarize the results on the CIFAR-10 and CIFAR-100 datasets in Table~\ref{Table_3}. Our proposed method outperforms other layer-wise training methods and is the only one comparable to end-to-end training. One possible reason for the performance gap between greedy layer-wise training and E2EBP is that the usable information for classification tends to saturate in the middle layers~\cite{sakamoto2024end}.

\begin{table}[h!]
 \centering
  \caption{Test accuracy (\%) on CIFAR-10 and CIFAR-100. The best performance is in bold and the second best performance is underlined.}\label{large_scale_dataset}
\renewcommand{\arraystretch}{1}
    \begin{tabular}{ccccccc}
    \toprule
    \multirow{3}{*}{Methods}&
    \multicolumn{3}{c}{CIFAR-10 }&\multicolumn{3}{c}{CIAR-100}\cr
    \cmidrule(lr){2-4} \cmidrule(lr){5-7}
    &VGG-11&VGG-16&ResNet-18&VGG-11&VGG-16&ResNet-18\cr
    \midrule
    Backprop (SGD)
    &\textbf{90.37}
    &\textbf{92.64}
    &\textbf{93.02}
    &\textbf{68.64}
    &\textbf{72.93}
    &\textbf{75.61}\cr
    
    \midrule
    Greedy LS
    &89.37
    &89.59
    &90.40
    &65.42
    &67.92
    &69.72\cr
    Greedy LE
    &89.23
    &89.56
    &90.26
    &66.14
    &68.22
    &70.03\cr

    HSIC bottleneck 
    &64.32
    &68.75
    &68.98
    &44.52
    &47.28
    &53.45\cr
    NIK 
    &63.69
    &65.52
    &67.32
    &43.94
    &46.10
    &52.75\cr
    \midrule
    Ours 
    &\underline{90.22}
    &\underline{92.21}
    &\underline{91.73}
    &\underline{67.63}
    &\underline{69.75}
    &\underline{72.03}\cr
    \bottomrule
    \end{tabular}
    \label{Table_3}
\end{table}

\subsubsection{Traffic Sign Recognition}
We evaluate our approach on two datasets: the Chinese Traffic Sign Recognition Database (CTSRD)\footnote{\url{https://nlpr.ia.ac.cn/pal/trafficdata/recognition.html}}~\cite{yang2015towards} and the German Traffic Sign Recognition Benchmark (GTSRB)\footnote{\url{https://benchmark.ini.rub.de/}}~\cite{stallkamp2011german}. The CTSRD dataset contains 6,164 traffic sign images across 58 categories. Each image is resized to $64 \times 64$, with $80\%$ of the data used for training and $20\%$ for testing. The GTSRB dataset includes 43 traffic sign classes, with 39,209 training images and 12,630 test images. This dataset features images captured under varying lighting conditions and with diverse, complex backgrounds. In both datasets, target objects typically occupy a significant portion of the image, with the bounding box of each object of interest averaging about $20\%$ of the image area. The baseline CNNs used for CTSRD and GTSRB are a 5-layer fully convolutional network and VGG-16, respectively.

To tackle this task, we propose a single CNN architecture with two output heads. The first head is a bounding box regression layer, which uses a linear regression output to predict the four coordinates of the traffic sign's bounding box. The second head is a classification layer equipped with a softmax function, which outputs the probabilities for each class, with the highest probability indicating the predicted class. This two-head design enables our network to simultaneously detect and classify traffic signs in a unified framework.

The learning objective of the $l$-th layer in traffic sign recognition becomes:
\begin{equation}\label{object_detection}
\mathcal{R}_{l}=\beta H\big(z_{l}\big)+\textmd{CE}(g(z_l);y)+\alpha \textmd{Bbox}(r(z_l);b),
\end{equation}
where $g(\cdot)$ represents the auxiliary object classifier, and $r(\cdot)$ denotes the auxiliary bounding box regressor. Bbox represents $l_1$ loss, $y$ is the ground truth category and $b$ is the target bounding box containing 4 coordinates. Each layer is jointly trained with an auxiliary category classifier and bounding box regressor with one fully connected layer. We set $\alpha=0.001$ for all baseline methods, and set $\beta=0.01$ for our proposed method and Greedy LE. We test our method and baseline methods for object detection task in terms of the object detection accuracy and mean intersection over union (mIOU). Tables~\ref{object_detection} and \ref{german_object_detection} suggest that greedy layer-wise training has the potential to exceed the performance of SGD in real-world applications. Again, our method is consistently better than Greedy LS and Greedy LE. The performance of HSIC Bottleneck and NIK is omitted due to their relatively poor results.


\begin{table}[ht]
\centering
\caption{Results on Chinese Traffic Sign Recognition Database (CTSRD).}
\begin{tabular}[t]{lcc}
\hline
\bf{Method}&\bf{Classification Accuracy} (\%)&\bf{mIOU} (\%)\\
\midrule
Backprop (SGD) &98.23&86.31\\
\midrule
Greedy LS &98.62&87.45\\
Greedy LE&98.68&88.56\\
\midrule
{\bf Ours}&\textbf{98.73}&\textbf{89.23}\\
\hline
\end{tabular}
\label{object_detection}
\end{table}%

\begin{table}[ht]
\centering
\caption{Results on German Traffic Sign Recognition Benchmark (GTSRB). The performance in terms of mIOU is omitted due to the absence of ground truth for bounding boxes.}
\begin{tabular}[t]{lc}
\hline
\bf{Method}&\bf{Classification Accuracy} (\%) \\
\midrule
Backprop (SGD) &97.38 \\
\midrule
Greedy LS &97.32 \\
Greedy LE&97.48 \\
\midrule
{\bf Ours}&\textbf{97.88} \\
\hline
\end{tabular}
\label{german_object_detection}
\end{table}%

The authors acknowledge the existence of more challenging datasets in this domain, such as Tsinghua-Tencent 100K~\cite{zhu2016traffic}, which feature a higher number of test images captured in wild environments. In these datasets, traffic signs are typically tiny, often occupying less than $1\%$ of the image. Adapting to such datasets requires more advanced architectures like YOLO and additional strategies. For instance, large images can be divided into smaller patches, which are then fed into a baseline network to produce patch-level object detection results. Considering that the primary focus of this paper is on greedy layer-wise training, and the mainstream literature in this area primarily evaluates simple architectures on benchmark datasets like MNIST or CIFAR-10, we believe our evaluations on CTSRD and GTSRB, along with the extension to a two-head CNN architecture, represent a significant advancement in this field. Extending our approach to YOLOv3 on the Tsinghua-Tencent 100K is planned as future work.

\section{Conclusions}
In this work, we systematically observed the flow of information inside popular convolutional neural networks by virtue of two mutual information matrices and a modified information plane. We then proposed a new strategy for greedy layer-wise training by incorporating a deterministic information bottleneck (DIB) coupled with the matrix-based R\'enyi's $\alpha$-order entropy as a key ingredient. Our Greedy DIB scales to large-scale data and popular deep architectures and performs better than state-of-the-art layer-wise training methods. Greedy DIB also outperforms SGD in traffic sign detection task which includes both regression and classification heads in a network. To the best of our knowledge, our results provide the first empirical evidence in real-world applications showing that greedy layer-wise training can outperform traditional end-to-end backpropagation.

\begin{appendix}

\section{Matrix-based R{\'e}nyi's $\alpha$-order entropy functional}\label{MI_estimator}

 Given $\{\mathbf{x}_{i}\}_{i=1}^{n}\in \mathcal{X}$, each $\mathbf{x}_i$ can be a real-valued scalar or vector, and the Gram matrix $K\in \mathbb{R}^{n\times n}$ computed as $K_{ij}=\kappa(\mathbf{x}_{i}, \mathbf{x}_{j})$, a matrix-based analogue to R{\'e}nyi's $\alpha$-entropy can be given by the following functional:

\begin{equation}\label{Renyi_entropy}
H_{\alpha}(A)=\frac{1}{1-\alpha}\log_2 \left(\tr (A^{\alpha})\right)=\frac{1}{1-\alpha}\log_{2}\left(\sum_{i=1}^{n}\lambda _{i}(A)^{\alpha}\right),
\end{equation}
where $\alpha\in (0,1)\cup(1,\infty)$. $A$ is the normalized version of $K$, i.e., $A=K/\tr (K)$. $\lambda _{i}(A)$ denotes the $i$-th eigenvalue of $A$.

Similarly, given $n$ pairs of samples $\{\mathbf{x}_{i}, \mathbf{y}_{i}\}_{i=1}^{n}$, the joint entropy between two random variable $\mathbf{x}$ and $\mathbf{y}$ can be defined as:

\begin{equation}\label{Renyi_joint_entropy}
H_{\alpha}(A,B)=H_{\alpha}\left(\frac{A \circ B}{\tr (A \circ B)}\right),
\end{equation}
where $A_{ij}=\kappa_{1}(\mathbf{x}_{i}, \mathbf{x}_{j})$ , $B_{ij}=\kappa_{2}(\mathbf{y}_{i}, \mathbf{y}_{j})$ and $A\circ B$  denotes the Hadamard product between the matrices $A$ and $B$. Same to~\cite{yu2021deep}, we set $\kappa$ as radial basis function (RBF) kernel $\kappa(\mathbf{x}_{i},\mathbf{x}_{j})=\exp(-\frac{\|\mathbf{x}_{i}-\mathbf{x}_{j}\|^{2}}{2\sigma ^{2}})$ to obtain the Gram matrices and set $\alpha=1.01$\footnote{The R{\'e}nyi's $\alpha$-order entropy reduces to Shannon entropy when $\alpha$ is approaching to 1. Thus, we chose $\alpha=1.01$ to approximate Shannon entropy for fair comparisons because most of existing IB approaches are based on Shannon entropy functional.}. The kernel width $\sigma$ is set by calculating the $k$ ($k=10$) nearest distances for each samples in the min-batch with size of $n$ and take the mean. Then we choose kernel width $\sigma$ of gram matrix $K$ ($n \times n$) as the average of mean values for all $n$ samples. Based on Eqs.~(\ref{Renyi_entropy}) and (\ref{Renyi_joint_entropy}), we can estimate mutual information as $I(\mathbf{x};\mathbf{y})=H(\mathbf{x})+H(\mathbf{y})-H(\mathbf{x},\mathbf{y})$.

\end{appendix}

\bibliography{strings}

\end{document}